\documentclass[10pt,twocolumn,letterpaper]{article}

\usepackage{iccv}
\usepackage{times}
\usepackage{epsfig}
\usepackage{graphicx}
\usepackage{amsmath}
\usepackage{amssymb}
\usepackage{array}
\usepackage{booktabs}
\usepackage{multirow}
\usepackage{caption, subcaption}

\usepackage[pagebackref=true,breaklinks=true,letterpaper=true,colorlinks,bookmarks=false]{hyperref}

\iccvfinalcopy 

\def\iccvPaperID{879} 

\ificcvfinal\fi

\newcommand\m[1]{\mathbf{#1}}

\begin{document}

\graphicspath{{Images/}}
\DeclareGraphicsExtensions{.png,.jpg,.pdf,.jpeg}

\title{{\boldmath $x$}R-EgoPose: Egocentric 3D Human Pose from an HMD Camera}

\author{$\text{Denis Tome}^{1,2}$, $\text{Patrick Peluse}^2$, $\text{Lourdes Agapito}^1$ and $\text{Hernan Badino}^2$\\
$^1\text{University College London}$ \quad $^2\text{Facebook Reality Lab}$\\
{\tt\small \{D.Tome, L.Agapito\}@cs.ucl.ac.uk} \quad {\tt\small \{patrick.peluse, hernan.badino\}@fb.com}
}

\twocolumn[{
\renewcommand\twocolumn[1][]{#1}
\maketitle
\begin{center}
  \vspace{-3mm}
  \includegraphics[width=0.9\linewidth]{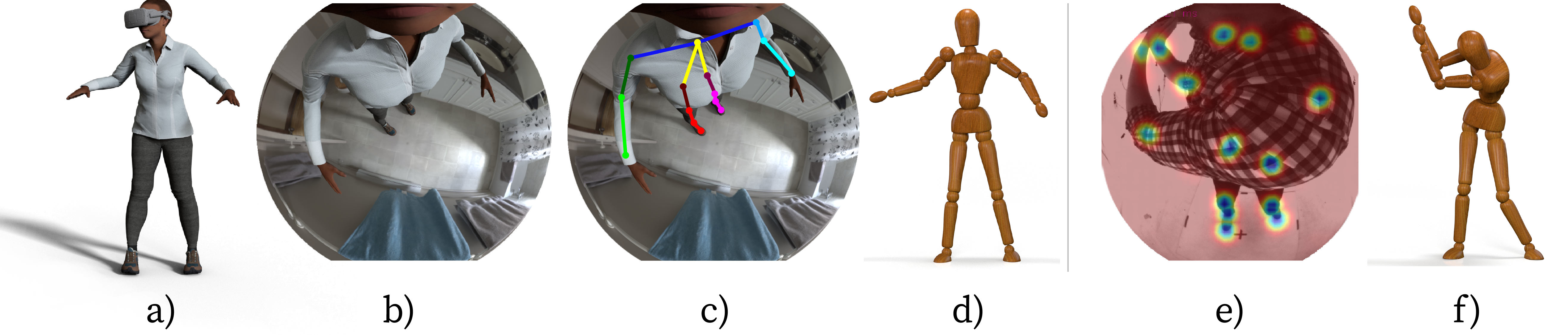} \vspace{-1mm}
  \captionof{figure}{\emph{\bf Left}: Our {\boldmath
      $x$}R-EgoPose~Dataset setup: (a) external camera viewpoint
    showing a synthetic character wearing the headset; (b) example of
    photorealistic image rendered from the egocentric camera
    perspective; (c) 2D and (d) 3D poses estimated with our
    algorithm. \emph{\bf Right}: results on real images; (e) real
    image acquired with our HMD-mounted camera with predicted 2D
    heatmaps; (f) estimated 3D pose, showing good generalization to
    real images.\label{fig:headset}} \vspace{-2mm}
\end{center}
}]


\begin{abstract}
  \vspace{-5mm} We present a new solution to egocentric 3D body pose
  estimation from monocular images captured from a downward looking
  fish-eye camera installed on the rim of a head mounted virtual
  reality device. This unusual viewpoint, just $2$ cm.~away from the
  user's face, leads to images with unique visual appearance,
  characterized by severe self-occlusions and strong perspective
  distortions that result in a drastic difference in resolution
  between lower and upper body. Our contribution is
  two-fold. Firstly, we propose a new encoder-decoder architecture
  with a novel dual branch decoder designed specifically to account
  for the varying uncertainty in the 2D joint locations. Our
  quantitative evaluation, both on synthetic and real-world datasets,
  shows that our strategy leads to substantial improvements in
  accuracy over state of the art egocentric pose estimation
  approaches. Our second contribution is a new large-scale
  photorealistic synthetic dataset -- {\boldmath $x$}R-EgoPose -- offering $383$K
  frames of high quality renderings of people with a diversity of skin
  tones, body shapes, clothing, in a variety of
  backgrounds and lighting conditions, performing a range of
  actions. Our experiments show that the high variability in our new
  synthetic training corpus leads to good generalization to real world
  footage and to state of the art results on real world datasets with
  ground truth. Moreover, an evaluation on the Human3.6M benchmark
  shows that the performance of our method is on par with top
  performing approaches on the more classic problem of 3D human pose
  from a third person viewpoint.
\end{abstract}

\section{Introduction}
The advent of {\boldmath $x$}R technologies (such as AR, VR, and MR)
have led to a wide variety of applications in areas such as
entertainment, communication, medicine, CAD design, art, and workspace
productivity. These technologies mainly focus on immersing the user
into a virtual space by the use of a head mounted display (HMD) which
renders the environment from the very specific viewpoint of the
user. However, current solutions have been focusing so far on the
video and audio aspects of the user's perceptual system, leaving a gap
in the touch and proprioception senses. Partial solutions to the
proprioception problem have been limited to hands whose positions are
tracked and rendered in real time by the use of controller
devices. The 3D pose of the rest of the body can be inferred from
inverse kinematics of the head and hand poses \cite{invkinavatars18},
but this often results in inaccurate estimates of the body
configuration with a large loss of signal which impedes compelling
social interaction \cite{uhess16} and even leads to motion sickness
\cite{reason1975motion}.

\begin{figure*}[!t]
  \vspace{-7mm}
  \centering
  \includegraphics[width=\linewidth]{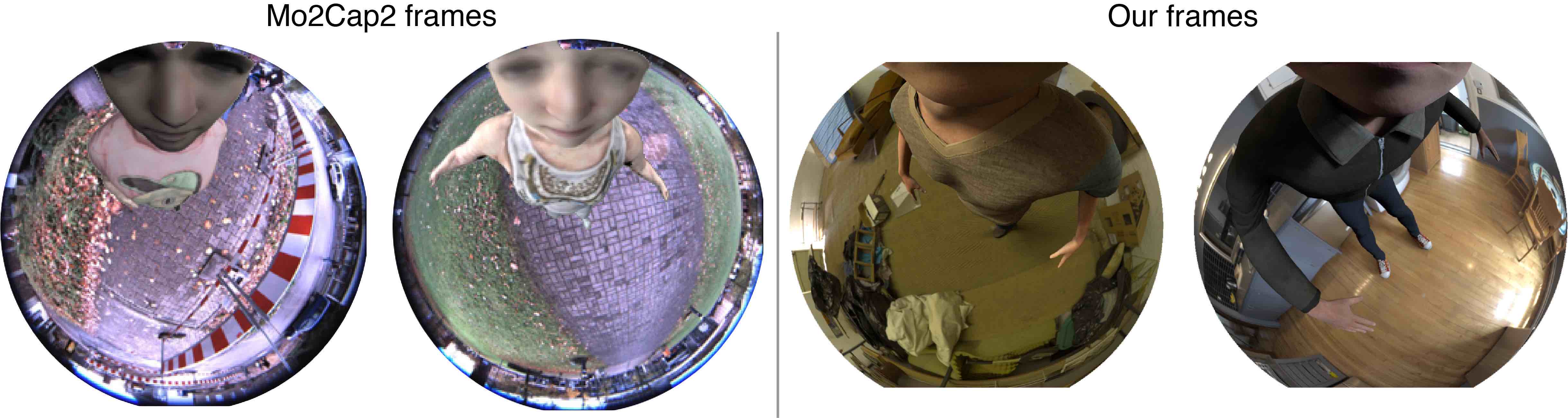}
  \vspace{-4mm}
  \caption{Example images from our {\boldmath $x$}R-EgoPose Dataset
    compared with the competitor Mo2Cap2
    dataset~\cite{xu2019mo2cap2}. The quality of our frames is far
    superior than the randomly sampled frames from mo2cap2, where the
    characters suffer color matching with respect to the background
    light conditions. \label{fig:comparison_mo2cap2}}
\vspace{-5mm}
\end{figure*}

In this paper we present a new approach for full-body 3D human pose
estimation \emph{from a monocular camera installed on a HMD}. In our
solution, the camera is mounted on the rim of a HMD looking down,
effectively just $2$cm.~away from an average size nose. With this
unique camera viewpoint, most of the lower body appears self-occluded
(see right images of Fig.~\ref{fig:comparison_mo2cap2}). In addition,
the strong perspective distortion, due to the fish-eye lens and the
camera being so close to the face, results in a drastic difference in
resolution between the upper and lower body (see
Fig.~\ref{fig:perspective}). Consequently, estimating 2D or 3D pose
from images captured from this first person viewpoint is considerably
more challenging than from the more standard external perspective and,
therefore, even state of the art approaches to human pose
estimation~\cite{rogez2018lrc-net++} underperform on our input data.

Our work tackles the two main challenges described above: \emph{(i)}
given the unique visual appearance of our input images and the
scarcity of training data for the specific scenario of a HMD mounted
camera, we have created a new large scale photorealistic synthetic
dataset for training with both 2D and 3D annotations; and \emph{(ii)}
to tackle the challenging problem of self-occlusions and difference in
resolution between lower and upper body we have proposed a new
architecture that accounts for the uncertainty in the estimation of
the 2D location of the body joints.

More specifically, our solution adopts a two-step approach. 
Instead of regressing directly the 3D pose from input images, we first
train a model to extract the 2D heatmaps of the body joints and then
regress the 3D pose via an auto-encoder with a dual branch
decoder. While one branch is trained to regress 3D pose from the
encoding, the other reconstructs input 2D heatmaps. In this way, the
latent vector is enforced to encode the uncertainty in the 2D joint
estimates.  The auto-encoder helps to infer accurate joint poses for
occluded body parts or those with high uncertainty. Both sub-steps are
first trained independently and finally end-to-end as the resulting
network is fully differentiable. The training is performed on real and
synthetic data. The synthetic dataset was created with a large variety
of body shapes, environments, and body motions, and will be made
publicly available to promote progress in this area.


Our contributions can be summarized as:
\begin{itemize}
\item A new encoder-decoder network for egocentric full-body 3D pose
  estimation from monocular images captured from a camera-equipped VR
  headset (Sec.~\ref{sec:auto_encoder}). Our quantitative evaluation
  on both synthetic and real-world benchmarks with ground truth 3D
  annotations shows that our approach outperforms previous state of
  the art~\cite{xu2019mo2cap2}. Our ablation studies show that the
  introduction of our novel decoder branch, trained to reconstruct the
  2D input heatmaps, is responsible for the drastic improvements in 3D
  pose estimation.
\item We show that our new approach generalizes, without
  modifications, to the standard scenario of an external front facing
  camera. Our method is currently the second best performing
  after~\cite{sun2018integral} on the Human3.6M benchmark.
\item A new large-scale training corpus, composed of $383$K frames,
  that will be made publicly available to promote progress in the area
  of egocentric human pose capture (see
  Section~\ref{sec:dataset}). Our new dataset departs from the only
  other existing monocular egocentric dataset from a headmounted
  fish-eye camera~\cite{xu2019mo2cap2} in its photorealistic quality
  (see Fig.~\ref{fig:comparison_mo2cap2}), different viewpoint (since
  the images are rendered from a camera located on a VR HMD), and its
  high variability in characters, backgrounds and actions.
\end{itemize}

\section{Related Work}
We describe related work on monocular (single-camera) marker-less 3D
human pose estimation focusing on two distinct capture setups:
\emph{outside-in} approaches where an external camera viewpoint is
used to capture one or more subjects from a distance -- the most
commonly used setup; and \emph{first person} or egocentric systems
where a head-mounted camera observes the own body of the user. While
our paper focuses on the second scenario, we build on recent advances
in CNN-based methods for human 3D pose estimation. We also describe
approaches that incorporate wearable sensors for first person human
pose estimation.

{\bf Monocular 3D Pose Estimation from an External Camera Viewpoint:}
the advent of convolutional neural networks and the availability of
large 2D and 3D training
datasets~\cite{ionescu2014human3,andriluka142d} has recently allowed
fast progress in monocular 3D pose estimation from RGB images captured
from external cameras. Two main trends have emerged: \emph{(i)} fully
supervised regression of 3D joint locations directly from images
~\cite{li20143D,park20163D,tekin2016structured,zhou2016deep,pavlakos2017coarse,mehta2017monocular}
and \emph{(ii)} pipeline approaches that decouple the problem into the
tasks of 2D joint detection followed by 3D
lifting~\cite{martinez2017simple,moreno20173D,ramakrishna2012reconstructing,akhter2015pose,zhou2017sparse,zhou2016sparseness,bogo2016keep,sanzari2016bayesian}.
Progress in fully supervised approaches and their ability to
generalize has been severely affected by the limited availability of
3D pose annotations for in-the-wild images. This has led to
significant efforts in creating photorealistic synthetic
datasets~\cite{rogez2016mocap,varol2017learning} aided by the recent
availability of parametric dense 3D models of the human body learned
from body scans~\cite{loper2015smpl}. On the other hand, the appeal of
two-step decoupled approaches comes from two main advantages: the
availability of high-quality off-the-shelf 2D joint
detectors~\cite{wei2016convolutional,newell2016stacked,pishchulin2016deepcut,cao2017realtime}
that only require easy-to-harvest 2D annotations, and the possibility
of training the 3D lifting step using 3D mocap datasets and their
ground truth projections without the need for 3D annotations for
images. Even simple architectures have been shown to solve this task
with a low error rate~\cite{martinez2017simple}. Recent advances are
due to combining the 2D and 3D tasks into a joint
estimation~\cite{rogez2017lcr,rogez2018lrc-net++} and using
weakly~\cite{wu2016single,tome2017lifting,tung2017adversarial,drover2018can,pavlakos2018learning}
or self-supervised losses~\cite{tung2017self,rhodin2018unsupervised}
or mixing 2D and 3D data for training~\cite{sun2018integral}.

{\bf First Person 3D Human Pose Estimation:} while capturing users
from an egocentric camera perspective for activity recognition has
received significant attention in recent
years~\cite{fathi2011understanding,ma2016going,cao2017egocentric},
most methods detect, at most, only upper body motion (hands, arms or
torso). Capturing full 3D body motion from head-mounted cameras is
considerably more challenging. Some head-mounted capture systems are
based on RGB-D input and reconstruct mostly hand, arm and torso
motions~\cite{rogez2015first,yonemoto2015egocentric}. Jiang and
Grauman~\cite{jiang2017seeing} reconstruct full body pose from footage
taken from a camera worn on the chest by estimating egomotion from the
observed scene, but their estimates lack accuracy and have high
uncertainty. A step towards dealing with large parts of the body not
being observable was proposed in~\cite{amer2018deep} but for external
camera viewpoints. Rhodin \emph{et al.}  \cite{rhodin2016egocap}
pioneered the first approach towards full-body capture from a
helmet-mounted stereo fish-eye camera pair. The cameras were placed
around 25 cm away from the user's head, using telescopic sticks, which
resulted in a fairly cumbersome setup for the user but with the
benefit of capturing large field of view images where most of the body
was in view.  Monocular head-mounted systems for full-body pose
estimation have more recently been demonstrated by Xu~\emph{et
  al.}~\cite{xu2019mo2cap2} (who propose a real-time compact setup
mounted on a baseball cap) although in this case the egocentric camera
is placed a few centimeters further from the user's forehead than in
our proposed approach. Our approach substantially outperforms
Xu~\emph{et al.}'s method~\cite{xu2019mo2cap2} by $20\%$ or more on
both indoor and outdoor sequences from their real world evaluation
dataset.

{\bf 3D Pose Estimation from Wearable Devices:} Inertial Measurement
Units (IMUs) worn by the subject provide a camera-free alternative
solution to first person human pose estimation. However, such systems
are intrusive and complex to calibrate. While reducing the number of
sensors leads to a less invasive configuration~\cite{von2017sparse}
recovering accurate human pose from sparse sensor readings becomes a
more challenging task. An alternative approach, introduced by
Shiratori \etal~\cite{shiratori2011motion} consists of a multi-camera
structure-from-motion (SFM) approach using 16 limb-mounted
cameras. Still very intrusive, this approach suffers from motion blur,
automatic white balancing, rolling shutter effects and motion in the
scene, making it impractical in realistic scenarios.

\section{Challenges in Egocentric Pose Estimation}
%

\begin{figure*}[t]
  \vspace{-5mm}
  \includegraphics[width=\linewidth]{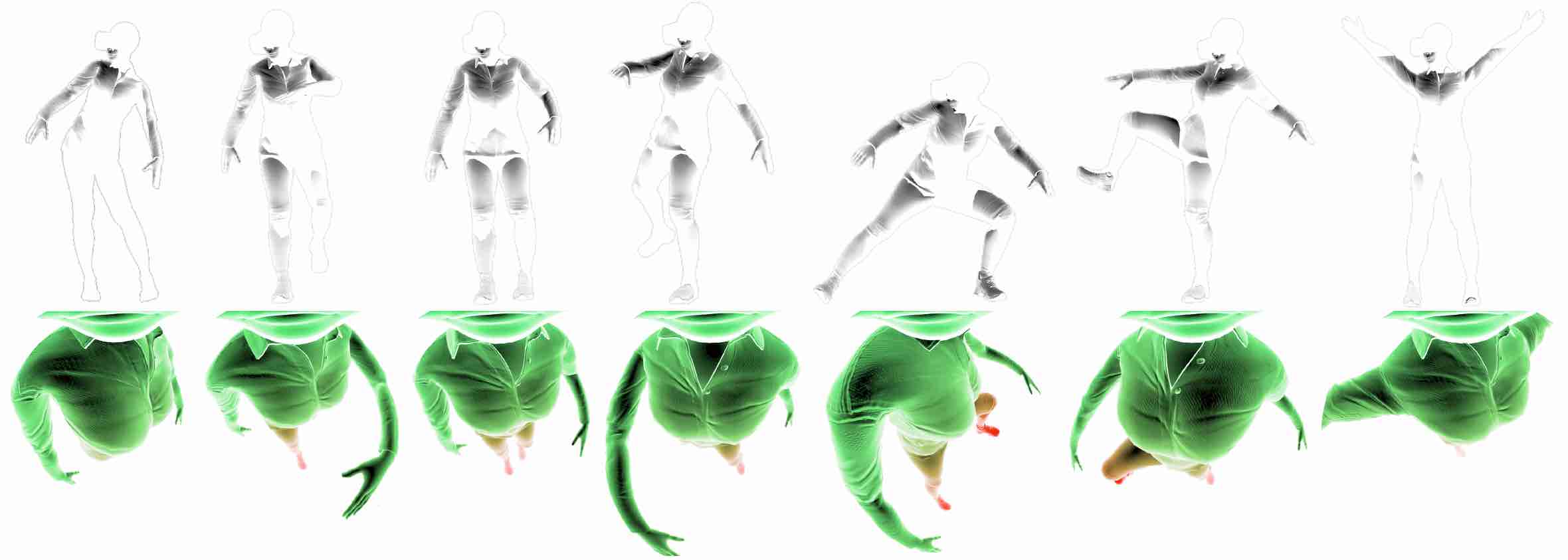}
  \vspace{-6mm}
  \caption{Visualization of different poses with the same synthetic
    actor. {\bf Top:} poses rendered from an external camera
    viewpoint. White represents occluded areas of the body.
    poses rendered from the egocentric camera viewpoint. The color
    gradient indicates the density of image pixels for each area of
    the body: \textit{green} indicates higher pixel density, whereas
    \textit{red} indicates lower density. This figure illustrates the
    most important challenges faced in egocentric human pose
    estimation: severe self-occlusions, extreme perspective effects
    and drastically lower pixel density for the lower
    body. \label{fig:perspective}}
\vspace{-4mm}
\end{figure*}



Fig.~\ref{fig:perspective} provides a visualization of the unique
visual appearance of our HMD egocentric setup --- the top row shows
which body parts would become self-occluded from an egocentric
viewpoint and dark green indicates highest and bright red lowest pixel
resolution. This unusual visual appearance calls both for a new
approach and a new training corpus. Our paper tackles both. Our new
neural network architecture is specifically designed to encode the
difference in uncertainty between upper and lower body joints caused
by the varying resolution, extreme perspective effects and
self-occlusions. On the other hand, our new large-scale synthetic
training set --- {\boldmath $x$}R-EgoPose --- contains $383$K images
rendered from a novel viewpoint: a fish-eye camera mounted on a VR
display. It has quite superior levels of photorealism in contrast with
the only other existing monocular egocentric
dataset~\cite{xu2019mo2cap2} (see Fig.~\ref{fig:comparison_mo2cap2}
for a side to side comparison), and large variability in the data. To
enable quantitative evaluations on real world images, we contribute
{\boldmath $x$}R-EgoPose$^\text{R}$, a smaller scale real-world
dataset acquired with a lightweight setup -- a real fish-eye camera
mounted on a VR display -- with ground truth 3D pose annotations. Our
extensive experimental evaluations show that our new approach
outperforms the current state of the art in monocular egocentric 3D
pose estimation~\cite{xu2019mo2cap2} both on synthetic and real-world
datasets.



\section{{\boldmath $x$}R-EgoPose Synthetic Dataset}
\label{sec:dataset}
The design of the dataset focuses on scalability, with augmentation of
characters, environments, and lighting conditions. A rendered scene is
generated from a random selection of characters, environments,
lighting rigs, and animation actions. The animations are obtained from
mocap data. A small random displacement is added to the positioning of
the camera on the headset to simulate the typical variation of the
pose of the headset with respect to the head when worn by the user.

\noindent\textbf{Characters}: To improve the diversity of body types,
from a single character we generate additional \textit{skinny short},
\textit{skinny tall}, \textit{full short}, and \textit{full tall}
versions The height distribution of each version varies from
$155$~cm.~to $189$~cm.

\noindent\textbf{Skin}: color tones include \textit{white} (Caucasian,
freckles or Albino), \textit{light-skinned European},
\textit{dark-skinned European} (darker Caucasian, European mix),
\textit{Mediterranean or olive} (Mediterranean, Asian, Hispanic,
Native American), \textit{dark brown} (Afro-American, Middle Eastern),
and \textit{black} (Afro-American, African, Middle
Eastern). Additionally, we built random skin tone parameters into the
shaders of each character used with the scene generator.

\noindent\textbf{Clothing}: Clothing types include Athletic Pants,
Jeans, Shorts, Dress Pants, Skirts, Jackets, T-Shirts, Long Sleeves,
and Tank Tops. Shoes include Sandals, Boots, Dress Shoes, Athletic
Shoes, Crocs. Each type is rendered with different texture and
colors. 

 \noindent\textbf{Actions}: the type of actions are listed in
 Table~\ref{tab:action_frames}.


\noindent\textbf{Images}: the images have a resolution of $1024 \times
1024$ pixels and 16-bit color depth. For training and testing, we
downsample the color depth to 8 bit. The frame rate is $30$
fps. \textit{RGB}, \textit{depth}, \textit{normals}, \textit{body
  segmentation}, and \textit{pixel world position} images are
generated for each frame, with the option for exposure control for
augmentation of lighting. Metadata is provided for each frame
including 3D joint positions, height of the character, environment,
camera pose, body segmentation, and animation rig.
 

\noindent\textbf{Render quality}: Maximizing the photorealism of the
synthetic dataset was our top priority. Therefore, we animated the
characters in Maya using actual mocap data \cite{Mixamo19}, and used a
standardized physically based rendering setup with V-Ray. The
characters were created with global custom shader settings applied
across clothing, skin, and lighting of environments for all rendered
scenes.

\subsection{Training, Test, and Validation Sets}
The dataset has a total size of $383$K frames, with $23$ male and $23$
female characters, divided into three sets: \emph{Train-set:} $252$K
frames; \emph{Test-set:} $115$K frames; and \emph{Validation-set:}
$16$K frames. The gender distribution is: \emph{Train-set:} 13M/11F,
\emph{Test-set:} 7M/5F and \emph{Validation-set:}
3M/3F. Table~\ref{tab:action_frames} provides a detailed description
of the partitioning of the dataset according to the different actions.
\begin{table}[htbp]
  \begin{center}
    \small
    \renewcommand{\arraystretch}{1.0}
    \begin{tabular}{lccc}
      \textbf{Action}  & \textbf{N. Frames} & \textbf{Size Train} & \textbf{Size Test} \\
      \toprule
      Gaming           & 24019              & 11153               & 4684               \\
      Gesticulating    & 21411              & 9866                & 4206               \\
      Greeting         & 8966               & 4188                & 1739               \\
      Lower Stretching & 82541              & 66165               & 43491              \\
      Patting          & 9615               & 4404                & 1898               \\
      Reacting         & 26629              & 12599               & 5104               \\
      Talking          & 13685              & 6215                & 2723               \\
      Upper Stretching & 162193             & 114446              & 46468              \\
      Walking          & 34989              & 24603               & 9971               \\
      \bottomrule
    \end{tabular}
  \end{center}
  \vspace{-5mm}
  \caption{Total number of frames per action and their distribution
    between train and test sets. Everything else not mentioned is
    validation data.\label{tab:action_frames}}
  \vspace{-5mm}
\end{table}

\begin{figure*}[t]
  \vspace{-7mm}
  \includegraphics[width=\linewidth]{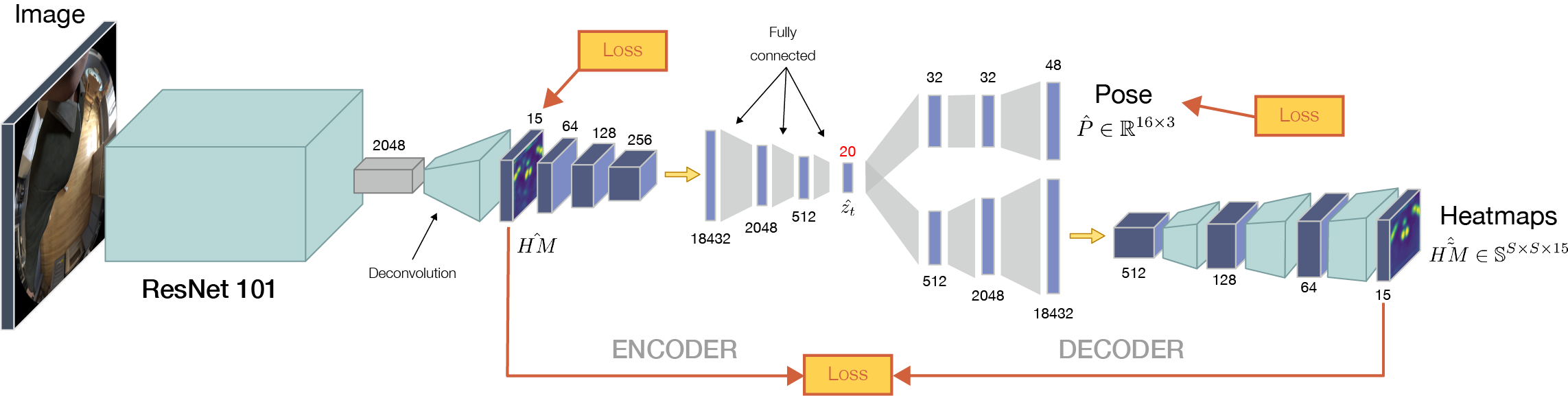}
  \vspace{-7mm}
  \caption{Our novel two-step architecture for egocentric 3D human
    pose estimation has two modules: \emph{a)} the 2D heatmap
    estimator, based on ResNet101~\cite{he2016deep} as the core
    architecture; \emph{b)} the 3D lifting module takes 2D heatmaps as
    input and is based on our novel dual branch
    auto-encoder.\label{fig:architecture}}
  \vspace{-5mm}
\end{figure*}

\section{Architecture}
\label{sec:model}
Our proposed architecture, shown in Fig.~\ref{fig:architecture}, is a
two step approach with two modules. The first module detects 2D
heatmaps of the locations of the body joints in image space using a
ResNet~\cite{he2016deep} architecture.  The second module takes the 2D
heatmaps as inputs and regresses the 3D coordinates of the body joints
using a novel dual branch auto-encoder.

One of the most important advantage of this pipeline approach is that
2D and 3D modules can be trained independently according to the
available training data. For instance, if a sufficiently large corpus
of images with 3D annotations is unavailable, the 3D lifting module
can be trained instead using 3D mocap data and projected heatmaps
without the need of paired images. Once the two modules are
pre-trained the entire architecture can be fine-tuned end-to-end since
it is fully differentiable. A further advantage of this architecture
is that the second branch is only needed at training time (see
Sec.~\ref{sec:auto_encoder}) and can be removed at test time,
guaranteeing the same performance and a faster execution.

\subsection{2D Pose Detection}
\label{sec:hm_regressor}
Given an RGB image $\m{I} \in \mathbb{R}^{368 \times 368 \times 3 }$
as input, the 2D pose detector infers 2D poses, represented as a set
of heatmaps $\m{HM} \in \mathbb{R}^{47 \times 47 \times 15}$, one for
each of the body joints. For this task we have used a standard
\textit{ResNet 101}~\cite{he2016deep} architecture, where the last
average pooling and fully connected layers have been replaced by a
deconvolutional layer, with $\text{kernel size}=3$ and
$\text{stride}=2$. The weights have been randomly initialized using
Xavier initialization~\cite{glorot2010understanding}. The model was
trained using normalized input images, obtained by subtracting the
mean value and dividing by the standard deviation, and using the mean
square error of the difference between the ground truth heatmaps and
the predicted ones as the loss:
\begin{equation}
  \text{L}_\text{2D} = \text{mse}(\m{HM}, \widehat{\m{HM}})
\end{equation}

We also trained alternative 2D pose detectors including the
CPM~\cite{wei2016convolutional} and the Stacked Hourglass
Network~\cite{newell2016stacked} resulting in comparable performance
at a higher computational cost.

\subsection{2D-to-3D Mapping}
\label{sec:auto_encoder}
The 3D pose module takes as input the $15$ heatmaps computed by the
previous module and outputs the final 3D pose ${\m{P} \in
  \mathbb{R}^{16\times3}}$. Note that the number of output 3D joints
is $16$ since we include the head, whose position cannot be estimated
in the 2D images, as the person is wearing a headset, but can be
regressed in 3D.
In most pipeline approaches the \textit{3D lifting} module typically
takes as input the 2D coordinates of the detected joints. Instead,
similarly to~\cite{pavlakos2018learning}, our approach regresses the
3D pose from heatmaps, not just 2D locations. The main advantage is
that these carry important information about the uncertainty of the 2D
pose estimates.

The main novelty of our architecture (see
Fig.~\ref{fig:architecture}), is that we ensure that this uncertainty
information is not lost. While the encoder takes as input a set of
heatmaps, and encodes them into the embedding $\hat{\m{z}}$, the
decoder has two branches -- one to regress the 3D pose from
$\hat{\m{z}}$ and another to reconstruct the input heatmaps. The
purpose of this branch is to force the latent vector to encode the
probability density function of the estimated 2D heatmaps.
%
%

The overall loss function for the auto-encoder becomes
\begin{eqnarray} 
  \text{L}_\text{AE} & = & \lambda_p(||\m{P} - \hat{\m{P}}||^2 + R(\m{P}, \hat{\m{P}})) + \nonumber \\   
 & & \lambda_{hm}||\widehat{\m{HM}} - \widetilde{\m{HM}}||^2 \label{eq:decoder}
 \end{eqnarray}
$\m{P}$ the ground truth; $\widetilde{\m{HM}}$ is the set of heatmaps
regressed by the decoder from the latent space and $\widehat{\m{HM}}$
are the heatmaps regressed by ResNet (see
Sec.~\ref{sec:hm_regressor}). Finally $R$ is the loss over the 3D
poses ${R(\m{P}, \hat{\m{P}}) = \lambda_\theta\theta(\m{P},
  \hat{\m{P}}) + \lambda_L\text{L}(\m{P}, \hat{\m{P}})}$ with
\begin{align*}
  & \theta(\m{P}, \hat{\m{P}}) = \sum_l^L\frac{\m{P}_l \cdot \hat{\m{P}}_l }{||\m{P}|| * ||\hat{\m{P}}_l||}
  & \text{L}(\m{P}, \hat{\m{P}}) = \sum_l^L||\m{P}_l-\hat{\m{P}}_l||
\end{align*}
corresponding to the cosine-similarity error and the limb-length
error, with $\m{P}_l \in \mathbb{R}^3$ the $l^{th}$ limb of the pose.
An important advantage of this loss is that the model can be trained
on a mix of 3D and 2D datasets simultaneously: if an image sample only
has 2D annotations then $\lambda_p=0$, such that only the heatmaps are
contributing to the loss.  In Section~\ref{sec:quantitative_eval} we
show how having a larger corpus of 2D annotations can be leveraged to
improve final 3D body pose estimates.

\begin{table*}[!t]
  \vspace{-3mm}
  \begin{center}
    \small
    \resizebox{\textwidth}{!}{
    \setlength\tabcolsep{3.0pt}
    \renewcommand{\arraystretch}{0.85}
    \begin{tabular}{lm{20mm}ccc >{\centering}m{15mm}ccc >{\centering}m{15mm}cc}
      \toprule
      Approach & Evaluation error (mm) & Gaming  & Gesticulating    & Greeting & Lower Stretching & 
       Patting & Reacting              & Talking & Upper Stretching & Walking  & All                                                                 \\
      \midrule
      \multirow{3}{*}{Martinez~\cite{martinez2017simple}}
               & Upper body            & 58.5    & 66.7             & 54.8     & 70.0             & 59.3  & 77.8  & 54.1  & 89.7  & 74.1  & 79.4     \\
               & Lower body            & 160.7   & 144.1            & 183.7    & 181.7            & 126.7 & 161.2 & 168.1 & 159.4 & 186.9 & 164.8    \\
               & Average               & 109.6   & 105.4            & 119.3    & 125.8            & 93.0  & 119.7 & 111.1 & 124.5 & 130.5 & 122.1    \\
      \midrule
      \multirow{3}{*}{\textbf{Ours - single-branch}} 
               & Upper body            & 114.4   & 106.7            & 99.3     & 90.9             & 99.1  & 147.5 & 95.1  & 119.0 & 104.3 & 112.5    \\
               & Lower body            & 162.2   & 110.2            & 101.2    & 175.6            & 136.6 & 203.6 & 91.9  & 139.9 & 159.0 & 148.3    \\
               & Average               & 138.3   & 108.5            & 100.3    & 133.3            & 117.8 & 175.6 & 93.5  & 129.0 & 131.9 & 130.4    \\
      \midrule
      \multirow{3}{*}{\textbf{Ours - dual-branch}} 
               & Upper body            & 48.8    & 50.0             & 43.0     & 36.8             & 48.6  & 56.4  & 42.8  & 49.3  & 43.2  & \bf 50.5 \\
               & Lower body            & 65.1    & 50.4             & 46.1     & 65.2             & 70.2  & 65.2  & 45.0  & 58.8  & 72.2  & \bf 65.9 \\
               & Average               & 56.0    & 50.2             & 44.6     & 51.1             & 59.4  & 60.8  & 43.9  & 53.9  & 57.7  & \bf 58.2 \\
      \bottomrule
    \end{tabular}}
  \end{center}
  \vspace{-5mm}
  \caption{ Quantitative evaluation with
    Martinez~\etal~\cite{martinez2017simple}, a state-of-the-art
    approach developed for front-facing cameras. Both upper and lower
    body reconstructions are shown as well. A comparison with our own
    architecture using a single-branch decoder is also included.  Note
    how the competing approach fails consistently across different
    actions in lower body reconstructions. This experiment emphasizes
    how, even a state-of-the-art 3D lifting method developed for
    external cameras fails on this challenging task.
    \label{tab:overall}}
  \vspace{-5mm}
\end{table*}

%
%
%

\subsection{Training Details}
The model has been trained on the entire training set for $3$ epochs,
with a learning rate of $1e-3$ using batch normalization on a
mini-batch of size $16$. The deconvolutional layer used to identify
the heatmaps from the features computed by \textit{ResNet} has
$\text{kernel size} = 3$ and $\text{stride} = 2$. The convolutional
and deconvolutional layers of the encoder have $\text{kernel size} =
4$ and $\text{stride} = 2$. Finally, all the layers of the encoder use
leakly ReLU as activation function with $0.2$ leakiness. The $\lambda$
weights used in the loss function were identified through grid search
and set to $\lambda_{hm}=10^{-3}$, $\lambda_{p}=10^{-1}$,
$\lambda_{\theta}=-10^{-2}$ and $\lambda_{L}=0.5$ . The model has been
trained from scratch with Xavier weight initializer.

\section{Quantitative Evaluation}
\label{sec:experiment}
We evaluate the proposed approach quantitatively on a variety of
egocentric 3D human pose datasets: \emph{(i)} the test-set of
{\boldmath $x$}R-EgoPose, our synthetic corpus, \emph{(ii)} the
test-set of {\boldmath $x$}R-EgoPose$^\text{R}$, our smaller scale
real-world dataset acquired with a real fish-eye camera mounted on a
VR display and with ground truth 3D poses, and \emph{(iii)} the
Mo2Cap2 test-set~\cite{xu2019mo2cap2} which includes 2.7K frames of
real images with ground truth 3D poses of two people captured in
indoor and outdoor scenes.

In addition we evaluate quantitatively on the Human3.6M dataset to
show that our architecture generalizes well without any modifications
to the case of an external camera viewpoint.

\noindent\textbf{Evaluation protocol}: Unless otherwise mentioned, we
report the Mean Per Joint Position Error - MPJPE:
\begin{equation}
  E(\m{P}, \hat{\m{P}}) = \frac{1}{N_f}\frac{1}{N_j}\sum_{f=1}^{N_f}\sum_{j=1}^{N_j}||\m{P}_{j}^{(f)}-\hat{\m{P}}_{j}^{(f)}||_2 \label{eq:eval_protocol}
\end{equation}
where $\m{P}_j^{(f)}$ and $\hat{\m{P}}_{j}^{(f)}$ are the 3D points of the
ground truth and predicted pose at frame $f$ for joint $j$,
out of $N_f$ number of frames and $N_j$ number of joints.

%
%
\subsection{Evaluation on our Egocentric Synthetic Dataset}
\label{sec:quantitative_eval}
\noindent\textbf{Evaluation on {\boldmath $x$}R-EgoPose test-set}:
Firstly, we evaluate our approach on the test-set of our synthetic
{\boldmath $x$}R-EgoPose dataset. It was not possible to establish a
comparison with state of the art monocular egocentric human pose
estimation methods such as Mo2Cap2~\cite{xu2019mo2cap2} given that
their code has not been made publicly available. Instead we compare
with Martinez \etal~\cite{martinez2017simple}, a recent state of the
art method for a traditional external camera viewpoint. For a fair
comparison the training-set of our {\boldmath $x$}R-EgoPose dataset
has been used to re-train this model. In this way we can directly
compare the performance of the 2D to 3D modules.


Table~\ref{tab:overall} reports the MPJPE (Eq.~\ref{eq:eval_protocol})
for both methods showing that our approach (Ours-dual-branch)
outperforms that by Martinez \etal by 36.4\% in the upper body
reconstruction, 60\% in the lower body reconstruction, and 52.3\%
overall, showing a considerable improvement.

\noindent\textbf{Effect of the second decoder branch:}
Table~\ref{tab:overall} also reports an ablation study to compare the
performance of two versions of our approach: with (Ours-dual-branch)
and without (Ours-single-branch) the second branch for the decoder
which reconstructs the heatmaps $\hat{\tilde{HM}}$ from the encoding
$\m{z}$. The overall average error of the single branch encoder is
$130.4$ mm, far from the $58.2$ mm error achieved by our novel
dual-branch architecture.

\begin{table*}[!t]
  \begin{center}
    \small
    \setlength\tabcolsep{3.0pt}
    \renewcommand{\arraystretch}{1.0}
    \begin{tabular}{l|cccccccccccccc}
      \toprule
      \textbf{Protocol \#1} & Chen                           & Hossain                       & Dabral                      & Tome                        & Moreno                      & Kanazawa               & Zhou 
                            & Jahangiri                      & Mehta                         & Martinez                    & Fang                        & Sun                         & Sun                    & \bf Ours \\
                            & \cite{chen20173d}              & \cite{hossain2018exploiting}* & \cite{dabral2017structure}* & \cite{tome2017lifting}      & \cite{moreno20173D}         & \cite{kanazawa2018end} & \cite{zhou2018monocap}
                            & \cite{jahangiri2017generating} & \cite{mehta2017monocular}     & \cite{martinez2017simple}   & \cite{fang2018learning}     & \cite{sun2017compositional} & \cite{sun2018integral} &          \\ 
      \midrule
      \textbf{Errors (mm)}  & 114.2                          & 51.9                          & 52.1                        & 88.4                        & 87.3                        & 88.0                   & 79.9
                            & 77.6                           & 72.9                          & 62.9                        & 60.4                        & 59.1                        & \bf 49.6               & 53.4     \\
      \midrule
      \midrule
      \textbf{Protocol \#2} & Yasin                          & Hossain                       & Dabral                      & Rogez                       & Chen                        & Moreno                 & Tome
                            & Zhou                           & Martinez                      & Kanazawa                    & Sun                         & Fang                        & Sun                    & \bf Ours \\
                            & \cite{yasin2016dual}           & \cite{hossain2018exploiting}* & \cite{dabral2017structure}* & \cite{rogez2016mocap}       & \cite{chen20173d}           & \cite{moreno20173D}    & \cite{tome2017lifting}
                            & \cite{zhou2018monocap}         & \cite{martinez2017simple}     & \cite{kanazawa2018end}      & \cite{sun2017compositional} & \cite{fang2018learning}     & \cite{sun2018integral}            \\
      \midrule
      \textbf{Errors (mm)}  & 108.3                          & 42.0                          & 36.3                        & 88.1                        & 82.7                        & 76.5                   & 70.7
                            & 55.3                           & 47.7                          & 58.8                        & 48.3                        & 45.7                        & \bf 40.6               & 45.24    \\      
    \end{tabular}
  \end{center}
 \vspace{-5mm}
  \caption{Comparison with other state-of-the-art approaches on the
    Human3.6M dataset (front-facing cameras). Approaches with * make
    use of temporal information.\label{tab:h36m}}
  \vspace{-5mm}
\end{table*}

\noindent\textbf{Reconstruction errors per joint type}:
Table~\ref{tab:joints_error} reports a decomposition of the
reconstruction error into different individual joint types. The
highest errors are in the hands (due to hard occlusions when they go
outside of the field of view) and feet (due to self-occlusions and low
resolution).

\subsection{Evaluation on Egocentric Real Datasets}
\label{sec:quantitative_eval}
\begin{table}[b]
  \vspace{-2mm}
  \begin{center}
    \small
    \renewcommand{\arraystretch}{1.0}
    \begin{tabular}{ll|ll}
      \textbf{Joint} & \textbf{Error (mm)} & \textbf{Joint} & \textbf{Error (mm)} \\
      \toprule
      Left Leg       & 34.33               & Right Leg      & 33.85               \\
      Left Knee      & 62.57               & Right Knee     & 61.36               \\
      Left Foot      & 70.08               & Right Foot     & 68.17               \\
      Left Toe       & 76.43               & Right Toe      & 71.94               \\
      Neck           & 6.57                & Head           & 23.20               \\
      Left Arm       & 31.36               & Right Arm      & 31.45               \\
      Left Elbow     & 60.89               & Right Elbow    & 50.13               \\
      Left Hand      & 90.43               & Right Hand     & 78.28               \\                  
    \end{tabular}
  \end{center}
  \vspace{-5mm}
  \caption{Average reconstruction error per joint using
    Eq.~\ref{eq:eval_protocol}, evaluated on the entire test-set (see
    Sec.~\ref{sec:dataset}). 
    \label{tab:joints_error}}
\end{table}
\begin{table*}[!t]
  \begin{center}
    \small
    \setlength\tabcolsep{9.0pt}
    \renewcommand{\arraystretch}{0.9}
    \begin{tabular}{lccccccccc}
      \toprule
       INDOOR                          & walking   & sitting   & crawling  & crouching & boxing    & dancing   & stretching & waving    & total (mm) \\
      \midrule
      3DV'17~\cite{mehta2017monocular} & 48.76     & 101.22    & 118.96    & 94.93     & 57.34     & 60.96     & 111.36     & 64.50     & 76.28      \\
      VCNet~\cite{mehta2017vnect}      & 65.28     & 129.59    & 133.08    & 120.39    & 78.43     & 82.46     & 153.17     & 83.91     & 97. 85     \\
      Xu~\cite{xu2019mo2cap2}          & 38.41     & 70.94     & 94.31     & 81.90     & 48.55     & 55.19     & 99.34      & 60.92     & 61.40      \\
      {\bf Ours}                       & \bf 38.39 & \bf 61.59 & \bf 69.53 & \bf 51.14 & \bf 37.67 & \bf 42.10 & \bf 58.32  & \bf 44.77 & \bf 48.16  \\
      \midrule
      \midrule
      OUTDOOR                          & walking   & sitting   & crawling  & crouching & boxing    & dancing   & stretching & waving    & total (mm) \\
      \midrule
      3DV'17~\cite{mehta2017monocular} & 68.67     & 114.87    & 113.23    & 118.55    & 95.29     & 72.99     & 114.48     & 72.41     & 94.46      \\
      VCNet~\cite{mehta2017vnect}      & 84.43     & 167.87    & 138.39    & 154.54    & 108.36    & 85.01     & 160.57     & 96.22     & 113.75     \\
      Xu~\cite{xu2019mo2cap2}          & 63.10     & \bf 85.48 & 96.63     & 92.88     & 96.01     & 68.35     & 123.56     & 61.42     & 80.64      \\
      {\bf Our}                        & \bf 43.60 & 85.91     & \bf 83.06 & \bf 69.23 & \bf 69.32 & \bf 45.40 & \bf 76.68  & \bf 51.38 & \bf 60.19  \\
      \bottomrule
    \end{tabular}
  \end{center}
 \vspace{-5mm}
  \caption{Quantitative evaluation on Mo2Cap2
    dataset~\cite{xu2019mo2cap2}, both indoor and outdoor
    test-sets. Our approach outperforms all competitors by more than
    \textbf{21.6\%} (13.24 mm) on indoor data and more than
    \textbf{25.4\%} (20.45 mm) on outdoor data.
    \label{tab:comp_mo2cap2}}
  \vspace{-3mm}
\end{table*}
\noindent\textbf{Comparison with Mo2Cap2~\cite{xu2019mo2cap2}}: We
compare the results of our approach with those given by our direct
competitor, Mo2Cap2, on their real world test set, including both
indoor and outdoor sequences. To guarantee a fair comparison, the
authors of~\cite{xu2019mo2cap2} provided us the heatmaps from their 2D
joint estimator. In this way, both 3D reconstruction networks use the
same input.  Table~\ref{tab:comp_mo2cap2} reports the MPJPE errors for
both methods. Our dual-branch approach substantially outperforms
Mo2Cap2~\cite{xu2019mo2cap2} in both indoor and outdoor
scenarios. Note that the dataset provided by the stereo egocentric
system EgoCap~\cite{rhodin2016egocap} cannot be directly used for
comparison, due to the hugely different camera position relative to
the head (their stereo cameras are 25cm.~from the head).

\begin{figure*}[tb]
  \includegraphics[width=\linewidth]{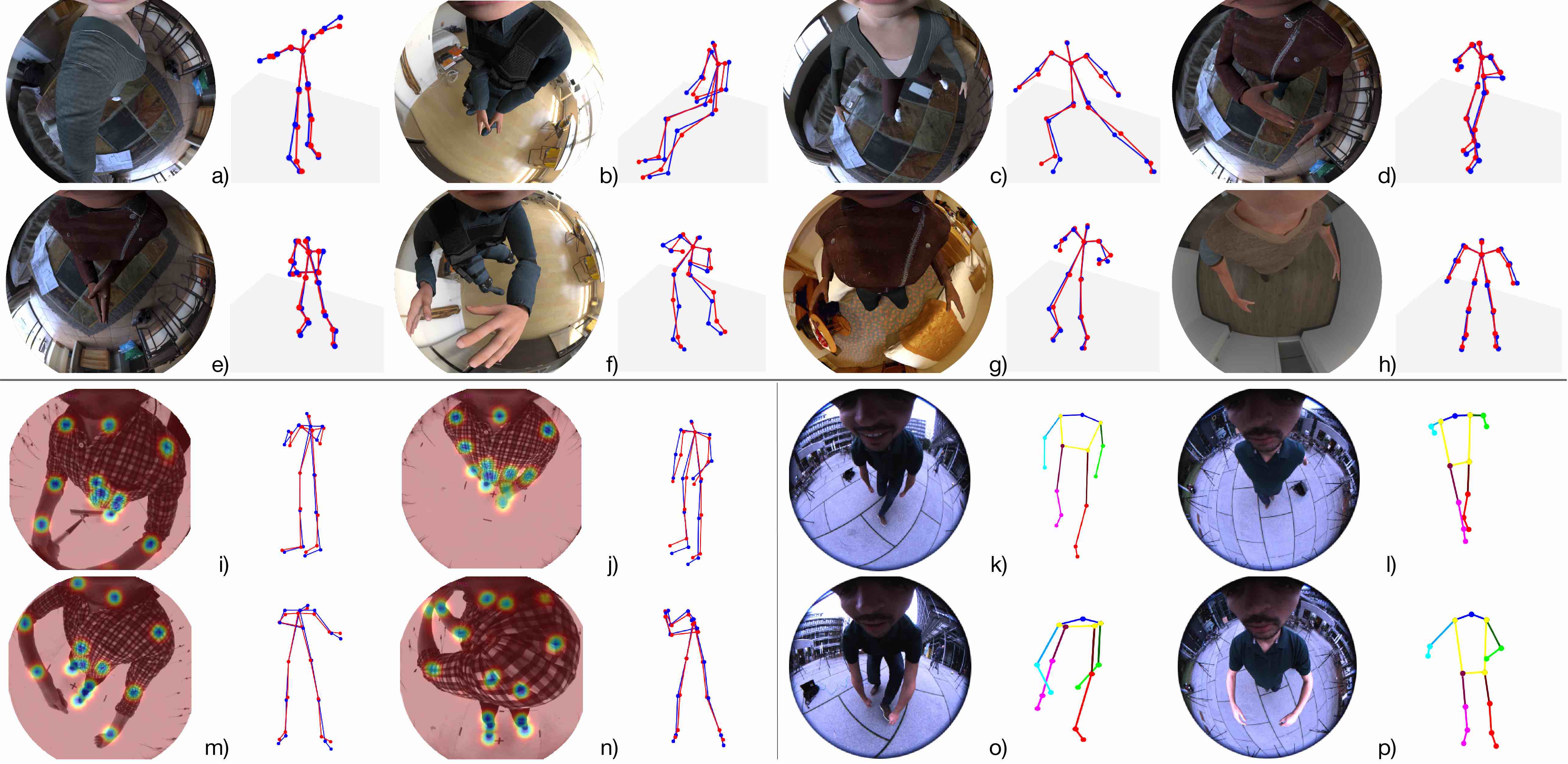}
  \vspace{-8mm}
  \caption{Qualitative results on synthetic and real images acquired
    with a camera physically mounted on a HMD: {\bf (top)} 3D poses
    reconstructed from synthetic images. Blue are ground truth poses
    and red predictions; {\bf (bottom)} reconstructed 3D predictions
    (in red) from \textbf{real images} captured in a mocap studio
    compared to ground truth poses (in blue), and reconstruction of
    images the wild from mo2cap2~\cite{xu2019mo2cap2} with poses shown
    using the same alignment for better
    visualization.\label{fig:qualitative_res}}
  \vspace{-5mm}
\end{figure*}

\noindent\textbf{Evaluation on {\boldmath $x$}R-EgoPose$^\text{R}$}:
The $\sim{}10$K frames of our small scale real-world data set were
captured from a fish-eye camera mounted on a VR HMD worn by three
different actors wearing different clothes, and performing 6 different
actions. The ground truth 3D poses were acquired using a custom mocap
system. The network was trained on our synthetic corpus ({\boldmath
  $x$}R-EgoPose) and fine-tuned using the data from two of the
actors. The test set contained data from the unseen third
actor. Examples of the input views and the reconstructed poses are
shown in Fig.~\ref{fig:qualitative_res}).  The
MPJPE~\cite{ionescu2014human3} errors (Eq.~\ref{eq:eval_protocol}) are
shown in Table~\ref{tab:real_res}. These results show good
generalization of the model (trained mostly on synthetic data) to real
images.

\begin{table}[!t]
  \begin{center}
    \small
    \renewcommand{\arraystretch}{1.0}
    \begin{tabular}{lc|lc}
      \textbf{Action}  & \textbf{Error (mm)} & \textbf{Action}  & \textbf{Error (mm)} \\
      \toprule
      Greeting         & 51.78 &      Upper Stretching & 61.09              \\
      Talking          & 47.46   &      Throwing Arrow   & 88.54             \\
      Playing Golf     & 68.74  &      \bf Average      & \bf 61.71                \\
      Shooting         & 52.64               \\
    \end{tabular}
  \end{center}
  \vspace{-5mm}
  \caption{Average reconstruction error per joint using
    Eq.~\ref{eq:eval_protocol}, evaluated on real data captured in a
    mocap studio.\label{tab:real_res}}
  \vspace{-7mm}
\end{table}
\subsection{Evaluation on Front-facing Cameras}
\noindent\textbf{Comparison on Human3.6M dataset}: We show that our
proposed approach is not specific for the egocentric case, but also
provides excellent results in the more standard case of front-facing
cameras. For this evaluation, we chose the Human3.6M dataset
~\cite{ionescu2014human3}.  We used two evaluation
protocols. \textit{Protocol 1} has five subjects (S1, S5, S6, S7, S8)
used in training, with subjects (S9, S11) used for evaluation. The
MPJPE error is computed on every 64th frame. \textit{Protocol 2}
contains six subjects (S1, S5, S6, S7, S8, S9) used for training, and
the evaluation is performed on every 64th frame of Subject 11
(Procrustes aligned MPJPE is used for evaluation).  The results are
shown in Table~\ref{tab:h36m}, from where it can be seen that our
approach is on par with state-of-the-art methods, scoring second
overall within the non-temporal methods.
%
%
\subsection{Mixing 2D and 3D Ground Truth Datasets}
An important advantage of our architecture is that the model can be
trained on a mix of 3D and 2D datasets simultaneously: if an image
sample only has 2D annotations but no 3D ground truth labels, the
sample can still be used and only the heatmaps will contribute to the
loss. We evaluated the effect of adding additional images with 2D but
no 3D labels on both scenarios: egocentric and front-facing
cameras. In the egocentric case we created two subsets of the
{\boldmath $x$}R-EgoPose test-set. The first subset contained $50$\%
of all the available image samples with both 3D and 2D labels. The
second contained $100$\% of the image samples with 2D labels, but only
50\% of the 3D labels. Effectively the second subset contained twice
the number of images with 2D annotations only.
Table~\ref{fig:quantitative:labels:ego_hmd} compares the results
between the two subsets, from where it can be seen that the final 3D
pose estimate benefits from additional 2D annotations. Equivalent
behavior is seen on the Human3.6M dataset. Figure
\ref{fig:quantitative:labels:h36m} shows the improvements in
reconstruction error when additional 2D annotations from COCO
\cite{lin2014microsoft} and MPII \cite{andriluka142d} are used.


\begin{table}[!b]
  \vspace{-8mm}
  \centering
  \small
  \resizebox{0.35\linewidth}{!}{
  \begin{subtable}[t]{0.4\linewidth}
    \centering
    \begin{tabular}{lll}
      3D & 2D           & Error (mm) \\
      \toprule
      50$\%$ & 50$\%$      & 68.04  \\
      50$\%$ & 100$\%$    & 63.98  \\
    \end{tabular}
    \caption{{\boldmath $x$}R-EgoPose}\label{fig:quantitative:labels:ego_hmd}
  \end{subtable}}
  \hfill
  \resizebox{0.55\linewidth}{!}{
  \begin{subtable}[t]{0.6\linewidth}
    \begin{tabular}{ll}
      Training dataset & Error (mm) \\
      \toprule
      H36M                & 67.9       \\
      H36M + COCO + MPII  & 53.4       \\
    \end{tabular}
    \caption{Human3.6M}\label{fig:quantitative:labels:h36m}
  \end{subtable}}
  \vspace{-4mm}
  \caption{Having a larger corpus of 2D annotations can be leveraged
    to improve final 3D pose
    estimation \label{tab:quantitative:labels}}
\end{table}
%
%
\begin{figure}[tb]
  \centering
  \includegraphics[width=\linewidth]{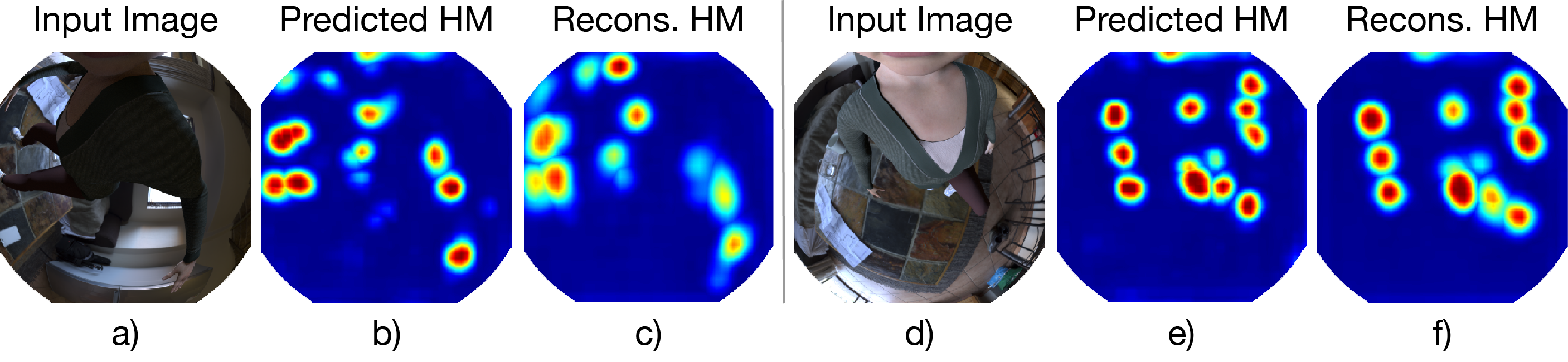}
  \vspace{-7mm}
  \caption{Examples of reconstructed heatmaps generated by the latent
    vector $\m{z}$. They reproduce the correct uncertainty of the
    predicted 2D joint positions.\label{fig:hm_reconstruction}}
  \vspace{-4mm}
\end{figure}

%
%
%
\subsection{Encoding Uncertainty in the Latent Space}
Figure~\ref{fig:hm_reconstruction} demonstrates the ability of our
approach to encode the uncertainty of the input 2D heatmaps in the
latent vector. Examples of input 2D heatmaps and those reconstructed
by the second branch of the decoder are shown for comparison.
%
%
%
\section{Conclusions}
We have presented a solution to the problem of 3D body pose estimation
from a monocular camera installed on a HMD. Our fully differentiable
network estimates input images to heatmaps, and from heatmaps to 3D
pose via a novel dual-branch auto-encoder which was fundamental for
accurate results.
We have also introduced the {\boldmath $x$}R-EgoPose dataset, a new
large scale photorealistic synthetic dataset that was essential for
training and will be made publicly available to promote research in
this exciting area. While our results are state-of-the-art, there are
a few failures cases due to extreme occlusion and the inability of the
system to measure hands when they are out of the field of view. Adding
additional cameras to cover more field of view and enable multi-view
sensing is the focus of our future work.


{\small
\bibliographystyle{ieee}
\bibliography{egbib}
}

\end{document}